\pdfoutput=1

\documentclass[11pt]{article}

\usepackage[]{acl}

\usepackage{times}
\usepackage{latexsym}
\usepackage{algorithm}
\usepackage{algorithmic}
\usepackage{graphicx}
\usepackage{booktabs}
\usepackage{color}
\newcommand{\secref}[1]{\S\ref{#1}}
\usepackage[T1]{fontenc}

\usepackage[utf8]{inputenc}

\usepackage{microtype}

\title{Autoregressive Entity Generation for End-to-End Task-Oriented Dialog}

\author{Guanhuan Huang, Xiaojun Quan \\
  School of Computer Science and Engineering\\
   Sun Yat-sen University, Guangzhou, China \\
  \texttt{huanggh25@mail2.sysu.edu.cn}\\
  \texttt{quanxj3@mail.sysu.edu.cn} \\\And
  Qifan Wang\\
  Meta AI\\
	Menlo Park, CA, USA\\
  \texttt{wqfcr@fb.com} \\}
\date{}

\begin{document}
\maketitle
\begin{abstract}
    Task-oriented dialog (TOD) systems often require interaction with an external knowledge base to retrieve necessary entity (e.g., restaurant) information to support the response generation. Most current end-to-end TOD systems either retrieve the KB information explicitly or embed it into model parameters for implicit access.~While the former approach demands scanning the KB at each turn of response generation, which is inefficient when the KB scales up, the latter approach shows higher flexibility and efficiency. In either approach, the systems may generate a response with conflicting entity information. To address this issue, we propose to generate the entity autoregressively first and leverage it to guide the response generation in an end-to-end system. To ensure entity consistency, we impose a trie constraint on entity generation. We also introduce a logit concatenation strategy to facilitate gradient backpropagation for end-to-end training. Experiments on MultiWOZ 2.1 single and CAMREST show that our system can generate more high-quality and entity-consistent responses.
\end{abstract}

\section{Introduction}

    Task-oriented dialog (TOD) systems \cite{young2013,budzianowski2018multiwoz} have become prominent and drawn much attention from both academia and industries. Their mission is to help users accomplish specific tasks such as booking restaurants and reserving hotels through natural language conversations, where an external knowledge base (KB) is usually needed to support the generation of a system response. For example, when trying to recommend a restaurant, they will retrieve its address from the KB and generate a response.

\begin{figure}[t]
		\centering
		\includegraphics[width=0.46\textwidth]{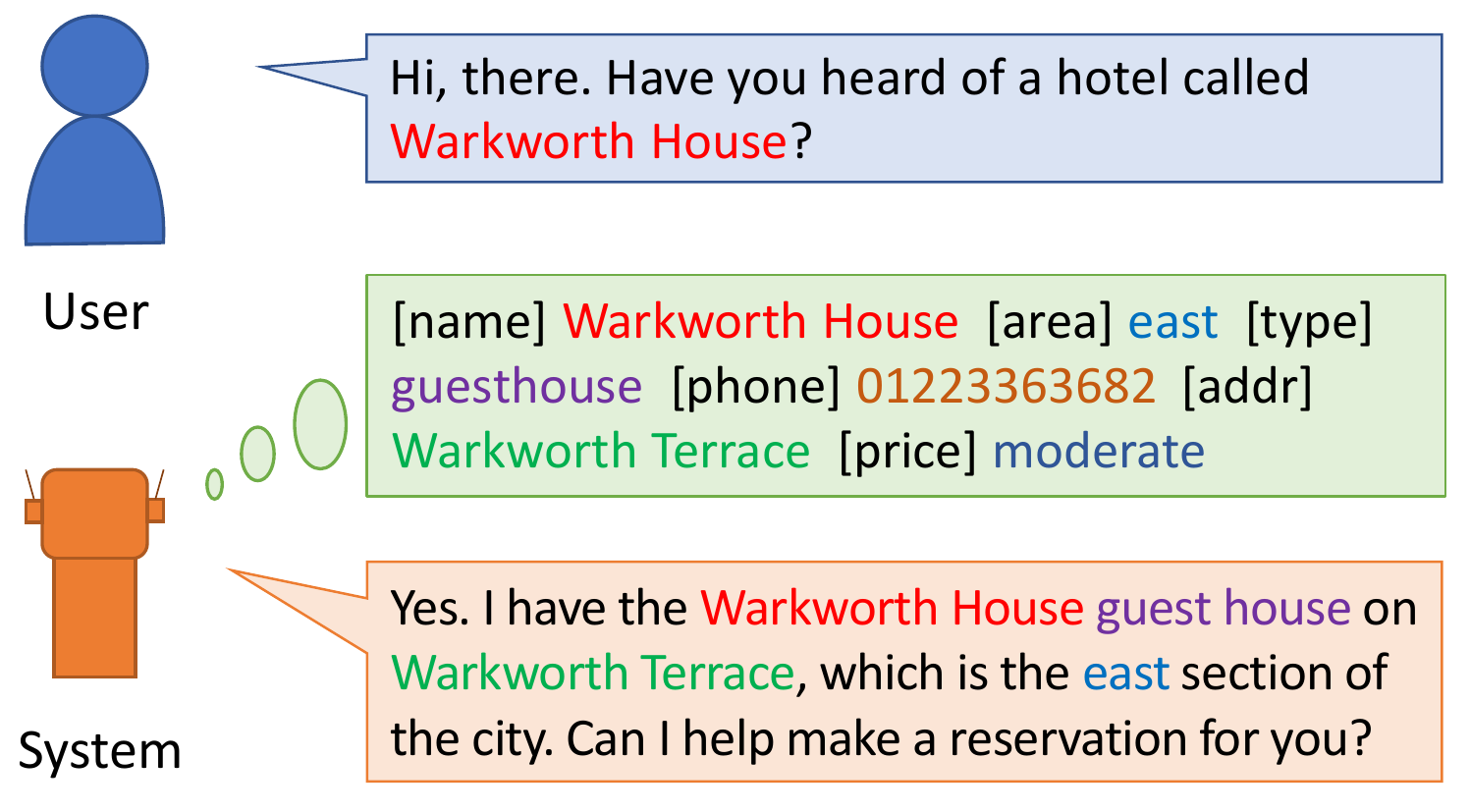}
		\caption{
			\label{fig:intro}
			An example to show that task-oriented dialog systems need to retrieve information (middle) from a knowledge base (KB) to generate a qualified system response. Entity values in the KB are color-highlighted. 
		}
\end{figure}

    Many recent state-of-the-art TOD systems \cite{mehri2019,hosseiniasl2020simpletod,Li2021} take a pipeline route that decomposes the task into modules that rely on intermediate annotations such as belief state and dialog act for supervision. These modules can be trained individually and assembled into a dialog system, mitigating the difficulty of generating a desired response directly from the dialog context and user utterance. Another motivation for the pipeline architecture is the necessity of querying KB with belief state, as shown in Figure \ref{fig:intro}, which would otherwise be non-trivial to realize in an end-to-end system. However, these annotations have to be crafted by human annotators, which is hardly realistic in practical scenes such as intelligent customer services where huge amounts of unannotated natural language conversations are accumulated. Besides, errors made in upstream modules may be propagated to downstream modules if they are not trained jointly.

    There are mainly two approaches to eliminating the reliance on intermediate annotations and generating system response in an end-to-end manner. First, entity information in the KB can be accessed by soft attention \cite{madotto2018mem2seq,reddy2019mlmn,qin2020dfnet}. To this end, a memory network is usually used to encode the KB, and attention and pointer are then utilized to retrieve entity information from the memory. These attention-based methods tend to become cumbersome when the KB scales up. Second, the KB information can be stored in model parameters to avoid direct interaction with the KB during response generation \cite{madotto2020gptke}. This is partly motivated by the observation that pre-trained models such as BERT \cite{devlin2018bert} can carry certain relational and factual knowledge \cite{petroni2019}.~To embed the KB into model parameters, this approach first augments the original training set with KB entries and then encodes the training samples with a powerful encoder. 

    Despite the success in end-to-end TOD systems, one of the remaining problems is entity inconsistency during response generation \cite{qin2019kbret}, which means that the systems usually generate conflicting entity information in system responses. For example, they may generate a response ``\texttt{Gourmet Kitchen is an Italian restaurant}'' while \texttt{Gourmet Kitchen} is actually a \texttt{North American} restaurant. In this work, we aim to address this issue more scalably in our end-to-end TOD system. Following GPT-KE \cite{madotto2020gptke}, we first insert the KB into natural language dialogs by data augmentation, so that the KB can be embedded into model parameters whose size does not scale with the KB. Then, we predict the entity that will appear in the response autoregressively. To avoid generating an inconsistent entity, we impose a trie constraint on the decoding to ensure that the generated entity truly belongs to the KB. The generated entity is taken as an extra input to generate an entity-consistent system response. Besides, since tokens in the entity are integers, which hinders gradient backpropagation, we propose logit concatenation to allow for end-to-end training.

    We evaluate our system on MultiWOZ 2.1 single \cite{budzianowski2018multiwoz} and CAMREST \cite{wen2017camr}, which are two task-oriented dialog benchmarks widely used in the literature. Experimental results show that it compares favorably with all the baselines. Particularly, it outperforms GPT-KE, a strong end-to-end TOD system that we follow, by a large margin. By ablation studies, we demonstrate that autoregressive entity generation assists in producing entity-consistent system responses in an end-to-end manner.

    To our best knowledge, this is the first work that attempts to alleviate the entity inconsistency problem in TOD systems by generating the entity first and taking it as an input for response generation. The system can be trained end-to-end without accessing external KBs during response generation.

\section{Related Work}
\label{sec:related works}
	
	End-to-end task-oriented dialog systems have drawn increasing attention in recent years. In one line of work, researchers propose to train the modules of a pipeline system jointly in an end-to-end framework, though they still require intermediate annotations for supervision. Among these works, SimpleTOD \cite{hosseiniasl2020simpletod}, SOLOIST \cite{peng2020soloist}, and UBAR \cite{yang2021ubar} attempt to concatenate the dialog history, user utterance, belief state, dialog act, and system response into a long sequence, which is then modeled by a sequence-to-sequence generation model. HyKnow \cite{gao2021hyknow} extends the belief state to handle both structured and unstructured knowledge and trains the dialog state tracking and response generation modules jointly. Nevertheless, these systems are not the end-to-end solutions we pursue in this work since they still need intermediate annotations. 
    
    There are mainly two approaches to implementing intermediate annotations free end-to-end TOD systems. First, entity information in the KB can be accessed by soft attention. Mem2Seq \cite{madotto2018mem2seq} combines the ideas of multi-hop attention over memory and a pointer network to incorporate KB information.  Wen et al. \shortcite{wen2018dsr} proposed to compute a dialogue state representation from the dialog history and use it to interact with KB representations to retrieve entity information for response generation. GLMP \cite{wu2019glmp} encodes the representations of dialog history and structural KB with a memory network and then passes the result to a decoder for response generation. DF-Net \cite{qin2020dfnet} includes a dynamic fusion module to generate a fused representation that explicitly captures the correlation between domains and uses it to query the KB. When the KB scales up, however, attention-based methods become less efficient. 
    
    Second, the KB information can be stored in model parameters to avoid further interaction with the KB during response generation. The motivation comes from the observation that pre-trained language models such as BERT \cite{devlin2018bert} and T5 \cite{raffed2020t5} can already carry certain relational and factual knowledge \cite{petroni2019}. GPT-KE \cite{madotto2020gptke} is the seminal dialog system towards this goal. It first augments the training set with KB entries and then learns a response generation model from the augmented set in an end-to-end fashion, thus abandoning the KB during response generation.

\section{Methodology}

	\begin{figure}[t]
		\centering
		\includegraphics[width=0.47\textwidth]{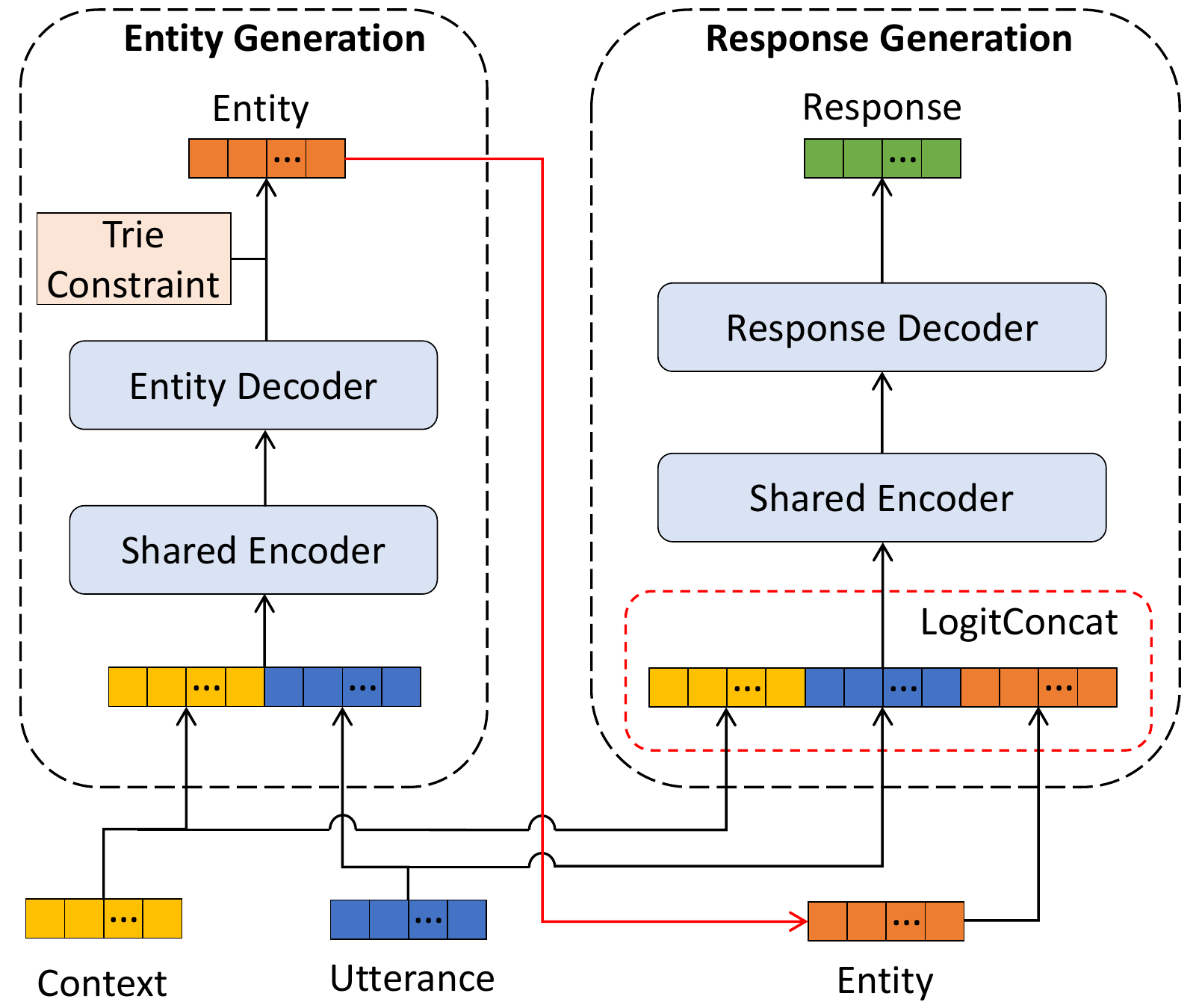}
		\caption{
			\label{fig:model}
			The architecture of our ECO system. The entity generation module takes the context and user utterance as input and generates a relevant entity. The response generation module takes the context, user utterance, and the generated entity as input to generate a system response. The two modules share the same encoder but have separate decoders. A trie constraint is imposed when generating the entity, and LogitConcat is proposed to facilitate end-to-end optimization.
		}
\end{figure}

    As shown in Figure \ref{fig:model}, our Entity-COnsistent end-to-end (ECO) task-oriented dialog system begins by embedding the KB into training dialogs (\secref{sec:preprocess}). Following GPT-KE \cite{madotto2020gptke}, we augment the original training set with KB entries and abandon the KB afterward. Unlike GPT-KE, which conducts data augmentation in data pre-processing before training, ECO conducts augmentation for each batch of training samples, which reduces the size of augmented training samples while maintaining high coverage of the KB. We then predict the entity (\secref{sec:auto_entity}) that will appear in the response and incorporate it into response generation (\secref{sec:supervised}) to ensure entity consistency, where LogitConcat is proposed to facilitate end-to-end optimization.

	

\subsection{Notations}
\label{sec:predefine}

	Given a training set $\mathcal{D}_{tr} = \left\{D_1, D_2, \dots, D_N\right\}$ of dialogs, where $D_i = \left\{U_{i,1}, R_{i,1}, \dots, U_{i,T}, R_{i,T}\right\}$ contains $T$ turns of user utterance and system response, we denote the conversational context of the $t$-th turn in dialog $D_i$ as $C_{i,t} = \left\{U_{i,1}, R_{i,1}, \dots, U_{i,t-1}, R_{i,t-1}\right\}$.~A structured knowledge base is given in the form of a set of entities $\mathit{KB} = \left\{E_1, E_2, \dots, E_M\right\}$, each of which is represented as a sequence  $E_i = \left\{a_1, v_{i,1}, a_2, v_{i,2}, \dots, a_K, v_{i,K}\right\}$ in which $a_j$ and $v_{i,j}$ denote the $j$th attribute and its value for entity $E_i$, respectively. For simplicity, we assume each turn of dialog only relates to one entity and reformulate it as $\left\{U_{i,t}, R_{i,t}, E_{i,t}\right\}$. A user goal \cite{schatzmann2007user_goal} is defined for each dialog as $G_i=(G_{i,c},G_{i,r})$, where $G_{i,c}$ specifies the constrained information (e.g., \{location=\texttt{center}, price=\texttt{cheap}\}) and $G_{i,r}$ denotes the required information (e.g., \texttt{address}, \texttt{name}).

\begin{figure}[t]
		\centering
		\fbox{
		\includegraphics[width=0.45\textwidth]{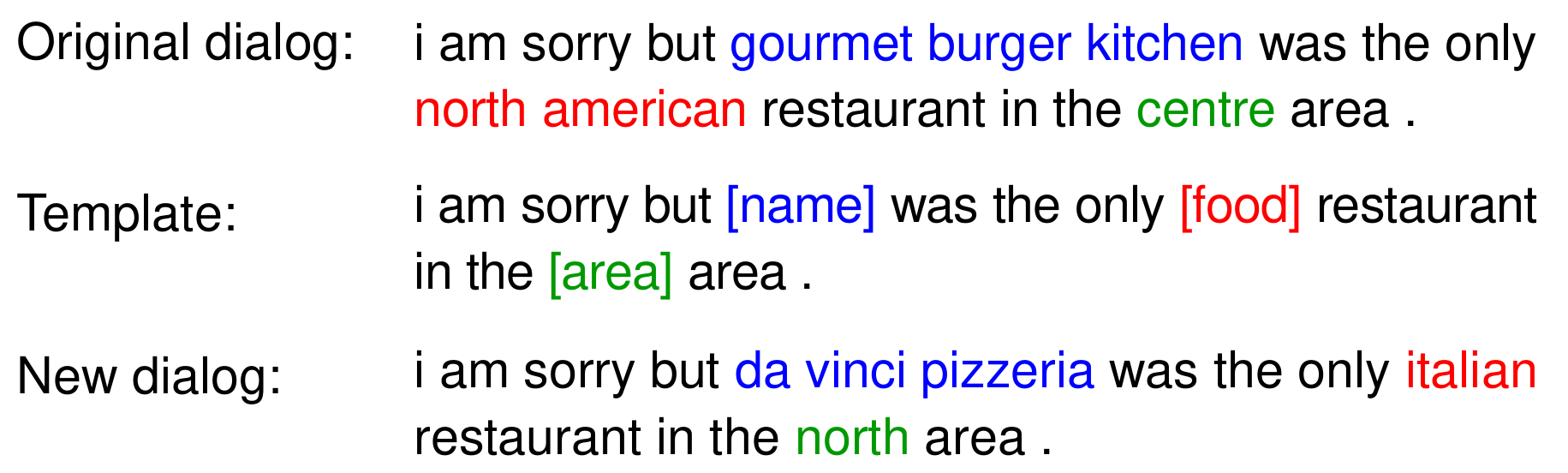}
		}
		\caption{
			\label{fig:knowledge_embed}
			An example to show how to construct a template from the training sample and generate a new sample from the template. Attributes are color-highlighted.
		}
\end{figure}	

\subsection{Knowledge Base Embedding}
\label{sec:preprocess}

	To embed the KB into the training set, we first extract all mentioned entity values in both user utterances and ground truth responses based on given span annotations in the original training set. Then, we match entity values with the KB to identify which entity is mentioned in the current turn of conversation. Templates are then constructed by replacing entity-related tokens in the conversation with special attribute placeholders.~For example, \texttt{north american} in Figure \ref{fig:knowledge_embed} is replaced with the corresponding attribute placeholder \texttt{[food]}. This template generation function is denoted as $\mathrm{DELEX}(\cdot)$, which is used to generate a set $\mathcal{D}_{tm}$ of templates from the original training set $\mathcal{D}_{tr}$:
    \begin{equation}
        \mathcal{D}_{tm}=\mathrm{DELEX}(\mathcal{D}_{tr}).
    \end{equation}

    Next, we generate new dialog samples with the templates in $\mathcal{D}_{tm}$. We refer to this process as data augmentation. To begin with, we obtain a set of KB entities, $\mathcal{G}_{mt} = \left\{E_1, E_2, \dots, E_{G}\right\}$, that match the predefined user goals.  Then, we randomly select an entity $E_i$ from $\mathcal{G}_{mt}$ and replace the placeholders with the corresponding values of $E_i$. For instance, we replace \texttt{[food]} and \texttt{[area]} in Figure \ref{fig:knowledge_embed} with \texttt{italian} and \texttt{north}, respectively. The function of generating samples from templates is defined as $\mathrm{RELEX}(\cdot)$, which is executed $P$ times to insert $P$ entities, producing a new set $\mathcal{D}_{au}$:
	\begin{equation}
	    \mathcal{D}_{au}=\bigcup_{p=1}^{P} \mathrm{RELEX}(\mathcal{D}_{tm}).
	\end{equation}
	
	Note that usually only a subset of samples in $\mathcal{D}_{tr}$ can successfully match with entities in KB during data augmentation, making $\mathcal{D}_{au}$ not cover all the samples of $\mathcal{D}_{tr}$. For this reason, we join $\mathcal{D}_{tr}$ and $\mathcal{D}_{au}$ to get our final training set $\mathcal{D}_{fn}$.
	
	\begin{equation}
	    \mathcal{D}_{fn}=\mathcal{D}_{tr} \bigcup \mathcal{D}_{au}
	\end{equation}
	
    The selected entities during the above augmentation process are treated as ground truth entities for the corresponding dialog samples.~This means that only the samples in $\mathcal{D}_{au}$ have entity labels while the samples in $\mathcal{D}_{tr}$ do not. Since all the placeholders in the templates are replaced with values from the same entity, this data augmentation process ensures that the augmented training samples contain consistent entity information.

\subsection{Autoregressive Entity Generation}
\label{sec:auto_entity}

    To predict the entity that will appear in the response, we propose to generate it autoregressively. For the sake of brevity, we use $C_t$ and $U_t$ to represent the current dialog context and user utterance, respectively. Then, we concatenate $C_t$ and $U_t$, encode them into a vector representation, and take it as an input for entity generation:
    \begin{equation}
        \label{eq:shared_encoder}
        \textbf{g}_{t}=\mathrm{Enc}(\mathrm{Emb}([C_t;U_t])),
    \end{equation}
    where $\mathrm{Emb}(\cdot)$ is the embedding function implemented by a global embedding matrix $\textbf{W}_e$.~$\mathrm{Enc}(\cdot)$ denotes the encoder which is shared with the response generation module (\secref{sec:supervised}).

    To generate an entity $\hat{E}_t$ autoregressively, the decoder iteratively predicts a token $\hat{e}_{t,k}$ based on the already generated sequence $\hat{E}_{t,<k}$ and vector representation $\textbf{g}_{t}$:
    \begin{equation}
        \label{eq:ent_dec}
        \hat{P}(\hat{e}_{t,k})=\mathrm{Dec}_{e}({\hat{e}}_{t,k}|\hat{E}_{t,<k},\textbf{g}_{t}).
    \end{equation}

    Since the gold entities of the samples in $\mathcal{D}_{au}$ are known, the cross-entropy loss of entity generation on $\mathcal{D}_{au}$ is defined as: 
    \begin{equation}
        \label{eqn:lentity}
        \mathcal{L}_{en} =  \sum_{D\in \mathcal{D}_{au}}\sum_{E_t \in D}\mathrm{CELoss}(\hat{E}_t,E_t),
    \end{equation}
    where $E_t$ denotes the ground truth entity for the $t$-th turn of conversation.

    For the samples in $\mathcal{D}_{tr}$, which have no entity labels, we do not calculate their loss during entity generation, but instead calculate their loss in response generation (\secref{sec:supervised}) to realize end-to-end optimization like DualTKB \cite{dognin2020dualtkb}.
    
\subsubsection{Trie Constraint}
\label{sec:tree}
	
\begin{figure}[]
		\centering
		\includegraphics[width=0.42\textwidth]{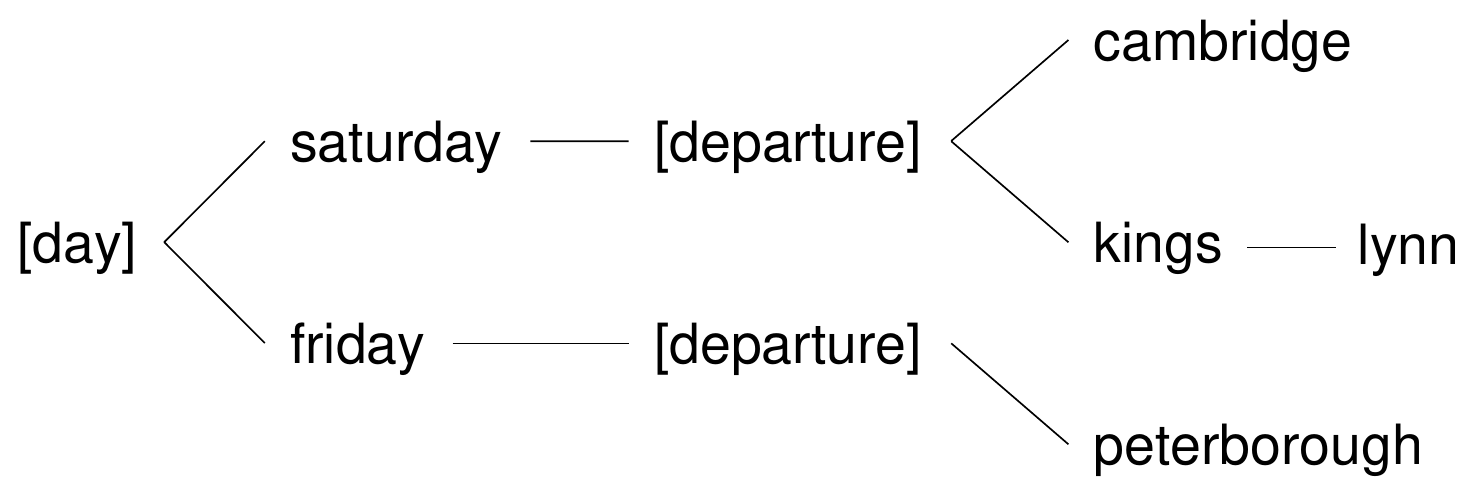}
		\caption{
			\label{fig:tree}
			A trie with three entity sequences: [\texttt{day}] \texttt{saturday} [\texttt{departure}] \texttt{cambridge}, [\texttt{day}] \texttt{saturday} [\texttt{departure}] \texttt{kings} \texttt{lynn}, and [\texttt{day}] \texttt{friday} [\texttt{departure}] \texttt{peterborough}.
		}
	\end{figure}

	Inspired by GENRE \cite{de2020autoregressive}, we construct a trie structure (a prefix tree) to ensure the generated entity truly belongs to the KB. For each entity in the KB, we construct a sequence as follows.~For each value in an entity, we put its attribute placeholder to precede it and concatenate all pairs of attribute and value to form a sequence such as \texttt{[name] cityroomz [area] centre [type] hotel}.~As depicted in Figure \ref{fig:tree}, a node in the trie denotes a token, and its child nodes denote all the succeeding tokens.
	
    When decoding the $k$-th token $\hat{e}_{t,k}$ during the generation of entity $\hat{E}_t$, we have the decoded sequence $\hat{E}_{t,<k}=\{\hat{e}_{t,1},\hat{e}_{t,2},\dots,\hat{e}_{t,k-1}\}$ in hand and walk through the trie along the path of $\hat{E}_{t,<k}$ to generate the next token. We use $\mathcal{E}_{t,k}$ to represent the set of possible tokens at this time step and re-compute $\hat{P}(\hat{e}_{t,k})$ as: 
	\begin{equation}
	    \label{eq:ent_gen_trie}
	    {P}(\hat{e}_{t,k})=\left\{
	        \begin{array}{ll}
		    \frac{\hat{P}(\hat{e}_{t,k})}{Z}, & \hat{e}_{t,k} \in \mathcal{E}_{t,k} \\
		    0, & \mathrm{else}  \\ 
	    \end{array}\right.
	\end{equation}
	where
	\begin{equation}
	    Z=\sum_{\hat{e}_{t,k} \in \mathcal{E}_{t,k}}\hat{P}(\hat{e}_{t,k}).
	\end{equation}

	Since only tokens from $\mathcal{E}_{t,k}$ have non-zero probabilities in ${P}(\hat{e}_{t,k})$, the model always samples a token from $\mathcal{E}_{t,k}$. Therefore, the generated entity is guaranteed to be valid constantly.
	
\subsection{Response Generation}
\label{sec:supervised}

    During training, for each sample in $\mathcal{D}_{au}$, the model generates a response based on the context, user utterance, and the corresponding ground truth entity by concatenating and encoding them with the shared encoder defined in Eq. (\ref{eq:shared_encoder}):
	\begin{equation}
	    \textbf{h}_t=\mathrm{Enc}(\mathrm{Emb}([C_t;U_t;E_t])).
	\end{equation}
	
	For each sample in $\mathcal{D}_{tr}$ which has no ground truth entity label, the generated entity is used:
	\begin{equation}
	\label{eq:dec_tr}
	   \textbf{h}_t=\mathrm{Enc}(\mathrm{Emb}([C_t;U_t;\hat{E}_t])).
	\end{equation}

	The response decoder then takes $\textbf{h}_t$ as input and generates the response $\hat{R}_t$ token by token as:
	\begin{equation}
	     P(\hat{r}_{t,k})=\mathrm{Dec}_r(\hat{r}_{t,k}|\hat{R}_{t,<k}, \textbf{h}_t).
	\end{equation}
	
    The cross-entropy loss is calculated between the generated response $\hat{R}_t$ and the ground truth response $R_t$:
    \begin{equation}
	    \label{eqn:lresp}
	    \mathcal{L}_{re}=\sum_{D\in \mathcal{D}^{}_{fn}}\sum_{R_t \in D}\mathrm{CELoss}(\hat{R}_t,R_t).
	\end{equation}
	
\subsubsection{Logit Concatenation}
\label{sec:LogitConcat}
	
	For those samples in $\mathcal{D}_{tr}$, since the tokens in each generated entity are integers, the gradients of response generation cannot be directly passed to the encoder during training. To address this, we modify Eq. (\ref{eq:dec_tr}) and input the distributions of generated entity tokens to the encoder. Specifically, for the $k$-th token $\hat{e}_{t,k}$ of a generated entity $\hat{E}_t$, its output distribution ${P}(\hat{e}_{t,k})$ over vocabulary from the entity decoder is first computed using Eq. (\ref{eq:ent_gen_trie}) and then used to approximate $\hat{e}_{t,k}$ for gradient propagation. ${P}(\hat{e}_{t,k})$ can be encoded as:
	\begin{equation}
	\label{eq:prob_emb}
	    \textbf{h}_{t,k}={P}(\hat{e}_{t,k})\textbf{W}_e^T,
	\end{equation}
    where $\textbf{W}_e$ is the global embedding matrix introduced above.

If both ${P}(\hat{e}_{t,k})$ and $\textbf{W}_e$ receive gradients during propagation, the training may collapse since it is much easier to update $\textbf{W}_e$ than ${P}(\hat{e}_{t,k})$, which requires understanding the context and utterance to obtain relevant information. Therefore, we alter the equation by stopping gradients on $\textbf{W}_e$:
	\begin{equation}
	    \hat{\textbf{h}}_{t,k}={P}(\hat{e}_{t,k})\cdot \mathrm{StopGrad}(\textbf{W}_e^T),
	\end{equation}
    \begin{equation}
        \hat{\textbf{h}}_{t}=\{\hat{\textbf{h}}_{t,1},\dots,\hat{\textbf{h}}_{t,|\hat{E}_t|}\}.
    \end{equation}

	We use $\hat{\textbf{h}}_{t}$ as the representation of entity $\hat{E}_t$ and concatenate it with the embeded context $C_t$ and user utterance $U_t$, which is then encoded to replace Eq. (\ref{eq:dec_tr}) during training:
	\begin{equation}
	    \textbf{h}_{t}=\mathrm{Enc}(\mathrm{Emb}([C_t;U_t]);\hat{\textbf{h}}_{t}]).
	\end{equation}

	Since ${P}(\hat{e}_{t,k})$ is a distribution vector rather than an integer, gradients can be backpropagated to the encoder during training.~At inference time, we take the generated entity tokens rather than ${P}(\hat{e}_{t,k})$ as input for response generation, as described in Eq. (\ref{eq:dec_tr}). This brings a gap between training and inference, which will be studied in Section \ref{sec:abla}. 
	
\subsection{Joint Training}
	
    The final system is trained by minimizing the sum of entity loss $\mathcal{L}_{en}$ and response loss $\mathcal{L}_{re}$:
	\begin{equation}
	    \mathcal{L}=\mathcal{L}_{en} + \mathcal{L}_{re}.
	\end{equation}

\section{Experiments}
\label{sec:experiment}

\begin{table*}[t]
		\centering
		\scalebox{0.83}{
			\begin{tabular}{lcccccc}
				\toprule
				\multicolumn{1}{l}{} & \textbf{BLEU}   & \textbf{Inform}  & \textbf{Success}   & \textbf{Score} & \textbf{F1} & \textbf{Consistency}      \\ 
				\midrule	
				Mem2Seq \cite{madotto2018mem2seq} & 6.60  & - & - & - & 21.62 & - \\
				DSR \cite{wen2018dsr}    & 9.10  & - & - & - & 30.00 & - \\
				GLMP \cite{wu2019glmp}   & 6.90  & - & - & - & 32.40 & -\\
				DF-Net \cite{qin2020dfnet}    & 9.40  & - & - & - & 35.10 & -\\
				GPT2 \cite{radford2019language}   & 14.33 & 64.60 & 51.77 & 72.52 & 30.38 & -\\
				GPT-KE \cite{madotto2020gptke} & \textbf{15.05} & 72.57 & 64.16 & 83.42 & 39.58 & 54.46\\
				\midrule
				BART-KE & 12.80$\pm$0.22 & 70.94$\pm$2.05 & 61.36$\pm$2.12 & 78.95$\pm$2.05 & 39.31$\pm$0.22 & 52.96$\pm$0.48\\
				ECO (ours)    & 12.61$\pm$0.20 & \textbf{83.63$\pm$0.63} & \textbf{75.37$\pm$0.21} & \textbf{92.11$\pm$0.20} & \textbf{40.87$\pm$0.24} & \textbf{56.84$\pm$0.36}\\ 
				\bottomrule
			\end{tabular}
		}
		\caption{
		\label{tab:main_woz} 
		Main results on MultiWOZ. Scores of baselines except BART-KE are from original papers, and ``-'' denotes scores originally unavailable. BART-KE is the baseline implemented by replacing GPT-2 in GPT-KE with BART.}
		
\end{table*}

\subsection{Dataset}

	We conduct experiments on MultiWOZ 2.1~single \cite{budzianowski2018multiwoz} and CAMREST \cite{wen2017camr}.~CAMREST consists of one domain of Cambridge restaurant booking while MultiWOZ 2.1 single consists of five domains:~\texttt{Attraction}, \texttt{Hotel}, \texttt{Restaurant}, \texttt{Taxi}, and \texttt{Train}.~Following previous work \cite{qin2020dfnet,madotto2020gptke}, we select only the dialogues which involves a single domain from MultiWOZ 2.1 to form the MultiWOZ 2.1 single dataset. We follow the same pre-processing and augmentation procedures as GPT-KE \cite{madotto2020gptke}.~Note that not all dialogs in the original training set can be successfully used to generate templates due to the diversity of entity values.~On MultiWOZ 2.1 single, 63/116/289/59 templates are respectively generated for domains \texttt{Attraction}/\texttt{Hotel}/\texttt{Restaurant}/\texttt{Train}, and no template is generated for the \texttt{Taxi} domain since MultiWOZ 2.1 single does not provide KB for this domain. On CAMREST, 161 templates are constructed for data augmentation.
	
	
	Following previous works \cite{qin2020dfnet,madotto2020gptke}, we adopt BLEU, Inform, Success, and F1 as the metrics to evaluate model performance on MultiWOZ 2.1 single, and employ BLEU, F1, and Success on CAMREST. Inform and Success are calculated based on the given user goal of a dialog session, and inconsistent entity information will lower the two metrics. Meanwhile, an overall score is also calculated: $\mbox{Score}=\mbox{BLEU}+(\mbox{Inform}+\mbox{Success})/2$.

\subsection{Measuring Entity Consistency}

    Measuring the entity consistency of a given system response remains a problem in task-oriented dialog systems. Qin et al. \shortcite{qin2021citod} annotated three kinds of inconsistency by human experts on the KVRET \cite{eric2017kvret} dataset, i.e., user query inconsistency, dialog history inconsistency, and knowledge base inconsistency. They then trained models as a form of automatic metrics to identify which kind of inconsistency appears in system responses. However, their method is trained on KVRET and may not be suitable to measure inconsistency on other datasets. 
Furthermore, the first few turns may include irrelevant entity information, such as providing several hotels for the user to choose, which makes it difficult to identify whether the generated response is dialog history consistent or not.   
    
    The above analysis motivates us to propose a new consistency metric that focuses on user query consistency and knowledge base consistency. It is a turn-level metric that requires all entity information in the user utterance and the system response to belong to the same entity in KB. To be specific, we first extract all entity information in the user utterance and the system response, and then search the knowledge base. If there is an entity that contains all the extracted information, this turn of conversation scores 1, and 0 otherwise. The final Consistency metric is calculated as the average score over all conversation turns. The method of extracting entity information from utterances and responses is the same as in calculating F1.
	
\subsection{Experiment Settings}

	Different from GPT-KE, we use BART \cite{lewis2020bart} as our backbone model due to the limitation of computation power. We also replace GPT-2 \cite{radford2019language} in GPT-KE with BART to form a new baseline, BART-KE. We set the max input sequence length to 256, the repeat times $P$ in $\mathrm{RELEX}$ to 12, and the batch size to 12. Experiments are conducted on a single NVIDIA 2080ti and cost about 11G GPU RAM. We conduct ablation studies on MultiWOZ 2.1 single as it is a more challenging dataset with multiple domains of dialogs. For most variants of our method, we run 30 epochs and evaluate them per 5 epochs, saving a model checkpoint after each evaluation. We then select the best checkpoint based on model performance on the development set and finally report the test results. For the ablation setting of \textit{w/ tr}, which lacks supervision from gold entity labels for training, we run 50 epochs to select the best.

\subsection{Main Results}
\label{sec:main_result}

	The overall results are shown in Table \ref{tab:main_woz} and Table \ref{tab:main_camr}. We observe that ECO outperforms GPT-KE and other baselines by a large margin in all metrics except BLEU, showing that ECO can reach the user goals of this dialog dataset more effectively. The improvement of ECO over BART-KE suggests that ECO's success mainly comes from the model design rather than BART itself. Specifically, by generating an entity with trie constraint to help response generation, ECO obtains consistent entity information and improves entity consistency of generated response. On the other hand, we note that BART-based methods (BART-KE and ECO) achieve relatively lower BLEU scores than the GPT-2 family baselines (GPT-2 and GPT-KE) on MultiWOZ. The main reason should be that we do not post-train BART with language modeling objectives on the training set, which affects the fluency of generated responses, while responses in MultiWOZ are more diverse across domains.
	
	\begin{table}[t]
        
		\centering
		\scalebox{0.8}{
			\begin{tabular}{lccc}
				\toprule
				\multicolumn{1}{l}{} & \textbf{BLEU}   & \textbf{F1}  & \textbf{Success}\\ 
				\midrule	
				KB-Trs & 14.80 & 45.30 & - \\
				MLMN & 13.61 & 54.85 & - \\
				BoSsNet & 15.20 & 43.10 & - \\
				KBRet & \textbf{18.64} & 55.76 & 62.03\\
				GPT-KE & 18.00 & 54.85 & 74.68 \\
				\midrule
				BART-KE & 17.84$\pm$0.28 & 70.42$\pm$0.37 & 75.06$\pm$1.52 \\
				ECO (ours) & 18.42$\pm$0.27 & \textbf{71.56$\pm$0.39} & \textbf{78.77$\pm$1.85}\\ 
				\bottomrule
			\end{tabular}
		}
		\caption{
		\label{tab:main_camr}
		Main results on CAMREST. KB-Trs \cite{haihong2019kbtrs}, MLMN \cite{reddy2019mlmn}, BoSsNet \cite{raghu2019bossnet}, KBRet \cite{qin2019kbret}, and GPT-KE \cite{madotto2020gptke} are baselines for comparison.
		}
		
\end{table}
	
\begin{table*}[t]
		\centering
		\scalebox{0.98}{
			\begin{tabular}{lcccc}
			    \toprule
			     & \textbf{Percentage (\%)} & \textbf{Inform} & \textbf{Success} & \textbf{F1}\\
			    \midrule
			    single inform & 46.0 & 91.67$\pm$1.63 & 83.01$\pm$1.98 & 61.09$\pm$0.81 \\
			    multi inform & 54.0 & 76.78$\pm$1.55 & 68.85$\pm$1.77 & 31.03$\pm$0.73 \\
			    single success & 84.5 & 83.60$\pm$1.23 & 77.49$\pm$0.43 & 42.74$\pm$0.10 \\
			    multi success & 15.5 & 83.81$\pm$2.69 & 63.81$\pm$1.35 &33.62$\pm$1.28 \\
			    total & 100.0 & 83.63$\pm$0.63 & 75.37$\pm$0.21 & 40.87$\pm$0.24 \\
			    \bottomrule
			\end{tabular}
		}
		\caption{
			\label{tab:multi_success}
			Results of study on how multiple matched entities affect evaluation metrics, where single/multi inform/success refer to the situation with single/multiple matched entities when calculating Inform/Success, and Percentage (\%) means the proportion of samples in the test set.
		}
\end{table*}

	
	
	
	We also analyze why the improvement of F1 is much smaller than that of Inform and Success on MultiWOZ. Inform and Success are calculated based on user goals, and in some circumstances, there are multiple entities that match a user goal. However, only the one that is mentioned in the ground truth response is counted as correct in F1. Therefore, a large improvement on Inform and Success means ECO achieves user goals better, but the entity mentioned in the generated response may be different from the one in the ground truth. As shown in Table \ref{tab:multi_success}, over 50\% of test samples have multiple matched entities when calculating Inform, and the percentage is 15.5\% when calculating Success. Multiple matched entities reduce model performance on all metrics, especially on F1. 
	
\begin{table*}[t]
		\centering
		\scalebox{0.93}{
			\begin{tabular}{lcccccc}
				\toprule
				\multicolumn{1}{l}{} & \textbf{BLEU}   & \textbf{Inform}  & \textbf{Success}   & \textbf{Score} & \textbf{F1} & \textbf{Consistency}\\ 
				\midrule		
				GPT-KE  & \textbf{15.05} & 72.57 & 64.16 & 83.42 & 39.58 & 54.46\\
				
				BART-KE & 12.80$\pm$0.22 & 70.94$\pm$2.05 & 61.36$\pm$2.12 & 78.95$\pm$2.05 & 39.31$\pm$0.22 & 52.96$\pm$0.48\\
				
				ECO & 12.61$\pm$0.20 & \textbf{83.63$\pm$0.63} & \textbf{75.37$\pm$0.21} & \textbf{92.11$\pm$0.20} & \textbf{40.87$\pm$0.24} & \textbf{56.84$\pm$0.36}\\
				
				\textit{\quad w/ au} & 8.94$\pm$0.06 & 79.20$\pm$2.26 & 56.34$\pm$0.55 & 76.71$\pm$0.94 & 30.38$\pm$1.67 & 55.49$\pm$0.33\\
				
				\textit{\quad w/ tr} & 11.21$\pm$0.37 & 67.55$\pm$4.41 & 54.87$\pm$4.38 & 72.42$\pm$4.07 & 36.45$\pm$1.17 & 52.43$\pm$1.89  \\
				
				\textit{\quad w/o trie}  & 12.40$\pm$0.36 & 80.68$\pm$0.91 & 72.12$\pm$1.25 & 88.80$\pm$0.81 & 39.81$\pm$0.25 & 56.31$\pm$0.62 \\
				\textit{\quad w/o LogitConcat} & 12.52$\pm$0.11 & 78.32$\pm$0.96 & 70.50$\pm$2.46 & 86.93$\pm$1.64 & 39.88$\pm$0.40 & 55.15$\pm$0.84   \\
				\textit{\quad w/ LogitEval} & 12.85$\pm$0.28 & 71.98$\pm$0.55 & 65.04$\pm$0.96 & 81.36$\pm$1.00 & 40.58$\pm$0.46 & 53.62$\pm$0.81 \\
				

				\bottomrule
			\end{tabular}
		}
		\caption{
			\label{tab:abla}
			Results of ablation studies. ECO \textit{w/ au} represents the ECO variant trained on samples of $\mathcal{D}_{au}$, and ECO \textit{w/ tr} represents the ECO variant trained on samples of $\mathcal{D}_{tr}$. ECO \textit{w/o trie} means that ECO does not apply the trie constraint during entity generation, while ECO \textit{w/o LogitConcat} means it does not apply LogitConcat during training. ECO \textit{w/o StopGrad} represents the ECO variant that drops $\mathrm{StopGrad}$ in LogitConcat. ECO \textit{w/ LogitEval} represents the ECO variant that applies LogitConcat during inference. 
		}
	\end{table*}
	
\begin{figure*}
	    \centering
	    \includegraphics[width=1\textwidth]{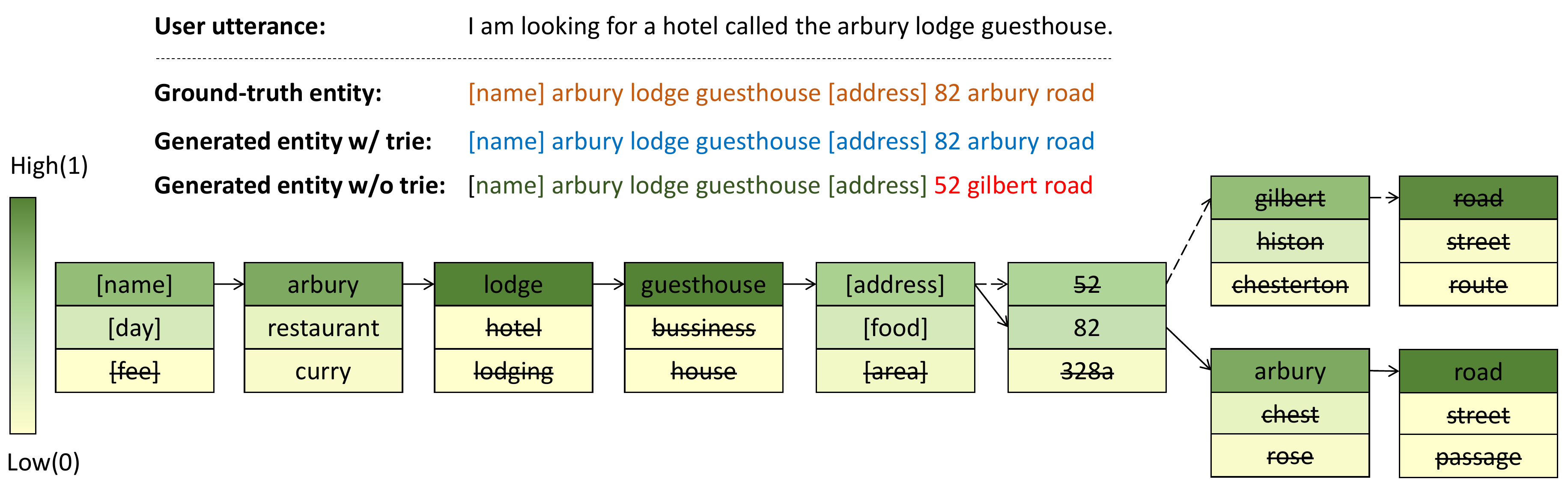}
	    \caption{An example of decoding entity with the trie constraint. Darker backgrounds represent high decoding probabilities. The trie constraint filters out some tokens during decoding and results in a different decoding path, avoiding the red path which has a higher probability but generates inconsistent entity information.} 
	    \label{fig:entity_heat}
\end{figure*}

\begin{figure*}[!htbp]
	    \centering
	    \includegraphics[width=0.65\textwidth]{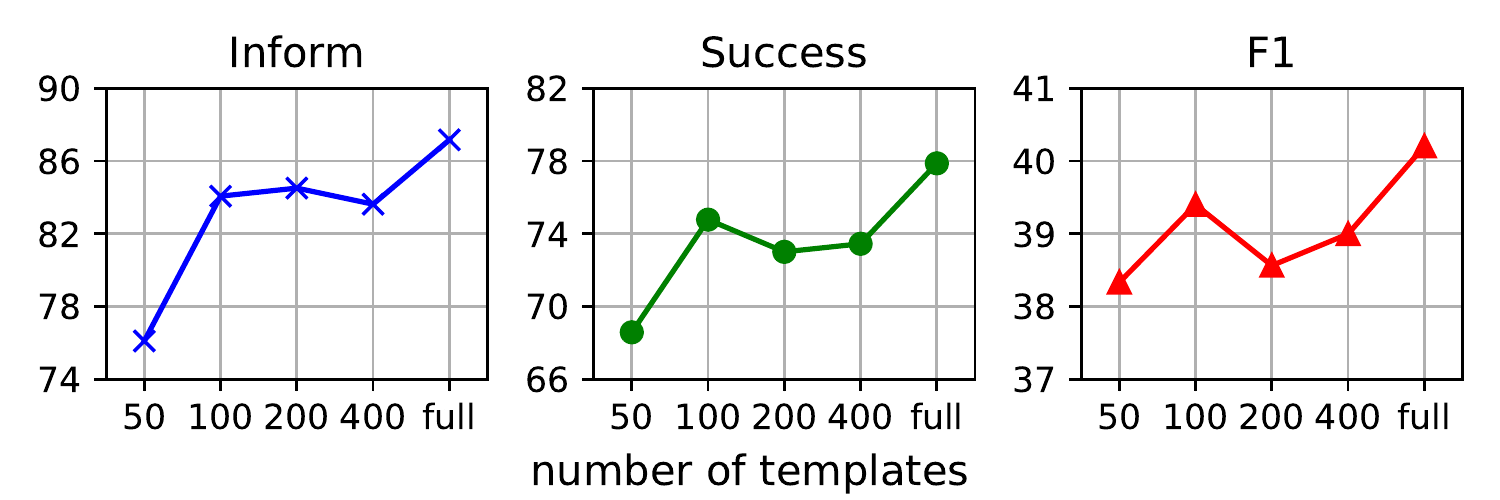}
	    \caption{Ablation study of how the number of templates affects model performance on MuitiWOZ, where \emph{full} means all the templates are used for knowledge base embedding.}
	    \label{fig:temp_num}
\end{figure*}
	
\subsection{Ablation Studies}
\label{sec:abla}

\subsubsection{Training Datasets}
	Unlike samples in $\mathcal{D}_{tr}$, samples in $\mathcal{D}_{au}$ have ground truth entity labels, so their training objectives are different. We use ECO \textit{w/ tr} to denote the training set that only contains samples from $\mathcal{D}_{tr}$ for end-to-end training, and use ECO \textit{w/ au} to denote the training set that only contains samples from $\mathcal{D}_{au}$. As the results show in Table \ref{tab:abla}, ECO \textit{w/ tr} has drops of 16.08 on Inform, 20.5 on Success, 4.42 on F1, and 4.41 on Consistency compared to ECO. Actually, without entity labels, the entity generation process is hard to converge, making response generation lack entity information as input and lowering model performance as a result.
	
On the other hand, ECO \textit{w/ au} has less drops than ECO \textit{w/ tr} compared to ECO on Inform, Success, and Consistency. However, the drops of ECO \textit{w/ au} is more obvious on F1, which is caused by the fact that $\mathcal{D}_{au}$ does not includes the \texttt{Taxi} domain. These results demonstrate that the augmentation introduces more important KB information for response generation than the original training set. 
	

\subsubsection{Trie Constraint}
	
In this work, the trie constraint is the key to guaranteeing entity consistency in a system response. Figure \ref{fig:entity_heat} presents an example of decoding an entity on the trie. Through filtering out non-kid nodes, the decoding path is restricted to a path on the trie. From Table \ref{tab:abla}, we note that ECO outperforms ECO \textit{w/o trie} by 2.95 on Inform, 3.25 on Success, 1.06 on F1, and 0.53 on Consistency. Thus, we can conclude that the model generates more informative responses with the help of consistently generated entities, which accounts for the improvement.

\subsubsection{Logit Concatenation}
	
LogitConcat is proposed to enable backpropagation after concatenating the dialogue context, user utterance, and the generated entity when training on $\mathcal{D}_{tr}$. Without this component, the model parameters of the entity generator will not be updated. In Table \ref{tab:abla}, ECO shows a promising improvement of 5.31 on Inform, 4.87 on Success, 0.99 on F1, and 1.69 on Consistency over ECO \textit{w/o LogitConcat}.

	

\subsubsection{Gap between Training and Evaluation}
	
During the evaluation, ECO uses the generated entity sequence as an input for response generation, which is different
from the training phase that uses LogitConcat. To study the impact, we conduct an experiment that also applies LogitConcat during the evaluation. From the results (ECO \emph{w/ LogitEval}) in Table \ref{tab:abla}, we observe obvious drops of ECO on Inform, Success, and Consistency. This phenomenon shows that applying LogitConcat during the evaluation weakens the consistency of the generated entities since the probability distribution in LogitConcat is not a valid entity from the KB.

\subsubsection{The Number of Templates}

To study how the number of templates affects model
performance, we randomly select several subsets
of templates from the whole template set. As shown in Figure
\ref{fig:temp_num}, the performance on all the metrics generally grows
when the number of templates increases, but there are fluctuations
when the number changes from 100 to 400.

\section{Conclusion}
\label{sec:conclusion}
	We proposed an end-to-end task-oriented dialog system by encoding external knowledge into model parameters. To address entity inconsistency in system responses, we proposed to generate the entities first and took them as input to response generation. To ensure the generated entities are valid, a trie constraint was imposed on the generation, and a logit concatenation strategy was introduced to facilitate backpropagation for end-to-end training. Experiments demonstrate that this system can produce more high-quality and entity-consistent responses in an end-to-end manner. For future work, we will extend this system to handle multiple entities that are involved in each turn of conversation.

\section*{Acknowledgments}
This work was supported by the National Natural Science Foundation of China (No. 62176270) and the Program for Guangdong Introducing Innovative and Entrepreneurial Teams (No. 2017ZT07X355).

\bibliography{anthology,custom}




\end{document}